\documentclass[journal]{IEEEtran}
\usepackage[english]{babel}
\usepackage{graphicx}
\usepackage{psfrag}
\usepackage{amsmath,amssymb}
\usepackage{subfigure}
\renewcommand{\vec}[1]{\hbox{\boldmath{$#1$}}}

\begin{document}
\title{Intelligent location of two simultaneously active acoustic emission sources: \\ Part II}

\author{Tadej Kosel and Igor Grabec\\
Faculty of Mechanical Engineering, University of Ljubljana,\\A\v sker\v ceva 6, POB 394, SI-1001 Ljubljana, Slovenia \\e-mail: tadej.kosel@guest.arnes.si; igor.grabec@fs.uni-lj.si
\thanks{Manuscript generated: January 31, 2007} }

\maketitle

\begin{abstract}
Part I describes an intelligent acoustic emission locator, while Part II
discusses blind source separation, time delay estimation and location of two continuous acoustic 
emission sources.

Acoustic emission (AE) analysis is used for characterization and
location of developing defects in materials. AE sources often generate a
mixture of various statistically independent signals. A difficult
problem of AE analysis is separation and characterization of signal
components when the signals from various sources and the mode of mixing
are unknown. Recently, blind source separation (BSS) by
independent component analysis (ICA) has been used to solve these
problems. The purpose of this paper is to demonstrate the applicability
of ICA to locate two independent simultaneously active acoustic emission sources
on an aluminum band specimen. The method is promising  for non-destructive testing of aircraft frame 
structures by acoustic emission analysis.  
\end{abstract}

\IEEEpeerreviewmaketitle

\section*{Introduction}

A common goal of many non-destructive testing methods is to detect
defects in materials. Acoustic emission analysis (AE) is a passive
testing method used to locate and characterize defects which
emit sound\cite{NTH}.

There are many ways to deduce the location of an AE source from
electrical signals detected by a chain of sensors. The corresponding
problems may be classified by the type of acoustic source mechanism as the
location of a continuous emission source, such as that generated by a leak,
or as the location of discrete emission, such as an AE burst caused by a growing crack.
This paper describes a method for processing continuous AE signals to
determine the time delay (T-D) between signals and thus to provide information for location of AE
sources. It should be pointed out that application of AE source characteristics, such as
count, count rate, amplitude distribution, and conventional time delay
measurement, becomes meaningless when dealing with continuous acoustic
sources.

The basic information for AE source location consists of T-D between
stress waves detected at different positions on a specimen. In the case of
only one active AE source, T-D of continuous acoustic waves can
be estimated using the cross-correlation function (CCF) of sensor
signals described in Part I of this article\cite{NTH,kosel00a}. In the case of two (or more)
simultaneously active AE sources, this method is not applicable, since
analysis of the CCF leads only to the T-D
of the most powerful AE signal. Detection of simultaneously active
independent AE source signals therefore requires a more sophisticated
approach.

The purpose of our study was to find a suitable method for processing
a mixture of two simultaneously active continuous AE signals to
determine the T-D and, related to this, the coordinates of both AE
sources. We found that the Blind Source Separation (BSS) method solves this
problem satisfactorily. BSS is a general signal processing method involving
the recovery of the contributions of different sources from a finite
set of observations recorded by sensors, independent of the propagation
medium and without any prior knowledge of the sources. BSS has
already been successfully applied in medicine, telecommunications, image
processing etc\cite{leebook98}. However, it is also a promising method for AE
analysis of aircraft structures, because AE signals are
often hidden in a mixture of signals from various sources. BSS could
extract the specific signature of each AE source, which can further be used
for location and characterization purposes, or to isolate AE sources from
background noise.
We conducted experiments with BSS on an aluminum beam on which two continuous AE 
sources were generated simultaneously by air flow.

\section*{Methods}
In this section we explain two different methods for time delay
estimation of AE sources. The first method is based on analysis of the
CCF and is convenient for T-D estimation
of one active continuous AE source as is described in Part I\cite{NTH,kosel00a,ziola91}.
The CCF exhibits a peak when the delay parameter
compensates the T-D between the sensor signals \cite{NTH}. The T-D is thus determined by
the position of the highest peak of the CCF. 
The second method is based on BSS algorithm and is convenient for T-D estimation
of two (or more) simultaneously active continuous AE sources\cite{lee97}.
Location of two simultaneously active AE
sources was performed by an intelligent locator based on a
general regression neural network\cite{grabecbook} as is described in Part I. 


Multichannel Blind Source Separation has recently received increased
attention due to the importance of its potential
applications\cite{burel92}. It occurs in many fields of engineering and
applied sciences, including processing of signals from antenna array, 
speech and geophysical data processing, noise reduction,
biological system analysis, etc. It consists of recovering signals
emitted by unknown sources and mixed by an unknown medium (material where waves propagate), using only
several observations of the mixtures. The only assumptions made are the
linearity of the mixing system and the statistical independence of
original signals.

BSS methods may be classified in several ways. One possible
classification that can be made depends on whether the mixtures are
instantaneous or convolutive \cite{deville97a}. Convolutive
mixtures correspond to a mixing system with time dependent memory. They
represent a more general case than instantaneous mixtures, and they have in
particular acoustic applications.  Recently, the principle
of independent component analysis (ICA) was applied in BSS, and it was
found to be a simple and powerful tool\cite{oja00}. This study deals with the
separation of two convolutively mixed independent continuous AE signals by ICA and 
the intelligent locator was used to locate two independent
continuous AE sources based on T-D \cite{grabecbook}.

The mixing and filtering processes of unknown input signals $s_j(t)$
may have different mathematical or physical backgrounds, depending on
specific applications. In this paper, we focus mainly on the simplest cases
with $n$ signals $x_i(t)$ linearly mixed in $n$ unknown statistically
independent, zero mean source signals $s_j(t)$. The composition is
expressed in matrix notation as  $\vec{x}=\vec{A} *
\vec{s}$ \cite{leebook98}, where `*' denotes a convolution,
$\vec{x}=[x_1(t),\ldots,x_n(t)]^{\text{T}}$ is the vector of sensor
signals, $\vec{s}=[s_1(t),\ldots,s_n(t)]^{\text{T}}$ is the vector of
source signals and $\vec{A}$ is an  unknown full rank $n \times n$
mixing matrix whose elements are finite inpulse response (FIR) filters. We assume that only vector
$\vec{x}$ is available. The goal of ICA is to find a matrix $\vec{W}$,
by which vector $\vec{x}$ can be transformed into source signals
$\vec{u}=\vec{W} * \vec{x}$.

Matrix $\vec{W}$ is simply the inverse of $\vec{A}$. However, when noise
corrupts the signals, matrix $\vec{W}$ must be found by an optimal
statistical treatment of the inverse problem. The optimal matrix
$\vec{W}$ can be estimated by a feed-forward neural network operating
in the frequency domain. A learning algorithm with Amari's
natural gradient can be written as\cite{amari98a}:
$\tilde{\vec{u}} =\tilde{\vec{W}} \cdot \tilde{\vec{x}}$, 
$\tilde{\vec{W}}(\tau+1) =\tilde{\vec{W}}(\tau)+\alpha\,\Delta
\tilde{\vec{W}}(\tau)+\eta\,\Delta \tilde{\vec{W}}(\tau-1)$, 
$\Delta \tilde{\vec{W}} = [\vec{I}-\tilde{\vec{y}}\cdot
\tilde{\vec{u}}^{\text{H}}]\,\tilde{\vec{W}}$, 
$\tilde{\vec{y}} =
\tanh(\Re[\tilde{\vec{u}}])+\imath\,\tanh(\Im[\tilde{\vec{u}}])$,
where $\alpha$ is the learning rate, $\eta$ is the constant of learning,
$\vec{I}$ is the identity matrix and the tilde `\~\,'represents a
frequency domain.

\begin{figure}[htb] 
\centering  
\psfrag{pred-procesiranje}[][]{\small pre-process}
\psfrag{inicializacija}[][]{\small initialize}
\psfrag{locitvenih}[][]{\small unmixing}
\psfrag{filtrov}[][]{\small filters}
\psfrag{FFT}[][]{\small FFT}
\psfrag{filter}[][]{\small filter}
\psfrag{tanh}[][]{\small tanh}
\psfrag{popravek locitvenih filtrov}[][]{\small update rule}
\psfrag{x(t)}[][]{\small $\vec{x}(t)$}
\psfrag{x}[][]{\small $\tilde{\vec{x}}$}
\psfrag{W}[][]{\small $\tilde{\vec{W}}$}
\psfrag{u}[][]{\small $\tilde{\vec{u}}$}
\psfrag{y}[][]{\small $\tilde{\vec{y}}$}
\psfrag{dW}[][]{\small $\Delta \tilde{\vec{W}}$}
\includegraphics{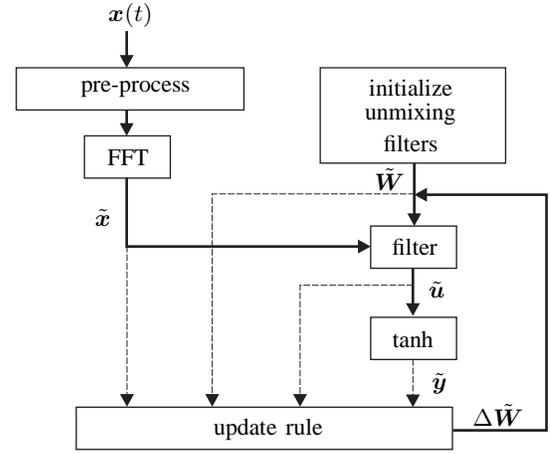}  
\caption{Block diagram of ICA algorithm} 
\label{ucno} 
\end{figure}  

The ICA algorithm runs off-line and proceeds as follows \cite{westner96} (Fig.~\ref{ucno}):
\begin{enumerate}
\setlength{\itemsep}{-3pt} 
\item Pre-process the time-domain input signals, $\vec{x}(t)$: substract the mean from each signal.
\item Initialize the frequency domain unmixing filters, $\tilde{\vec{W}}$.
\item Take a block of input data and convert it into the frequency domain using the Fast Fourier Transform (FFT).
\item Filter the frequency domain input block, $\tilde{\vec{x}}$, through $\tilde{\vec{W}}$ 
to get the estimated source signals, $\tilde{\vec{u}}$.
\item  Pass $\tilde{\vec{u}}$ through the frequency domain nonlinearity, $\tilde{\vec{y}}$.
\item Use $\tilde{\vec{W}}$, $\tilde{\vec{u}}$ and $\tilde{\vec{y}}$ along with the natural 
gradient extension \cite{amari96a} to compute the change in the unmixing filter, $\Delta \tilde{\vec{W}}$.
\item Take the next block of input data, covert it into the frequency domain, and proceed from step 4. 
Repeat this process until the unmixing filters have converged upon a solution, passing several times 
through the data.
\item Normalize $\tilde{\vec{W}}$ and convert it back into the time domain, using the Inverse Fast 
Fourier Transform (IFFT).
\item Convolve the time domain unmixing filters, $\vec{W}$, with $\vec{x}$ to get the estimated sources.
\end{enumerate}

\section*{Experiments}
\begin{figure}[htb] 
\centering  
\psfrag{fi2mm}[][]{\small $\phi$ 2\,mm}
\psfrag{trak}[][]{\small band}
\psfrag{x}[][]{\small $l$}
\psfrag{zracni curek}[][]{\small air flow}
\includegraphics[width=9cm]{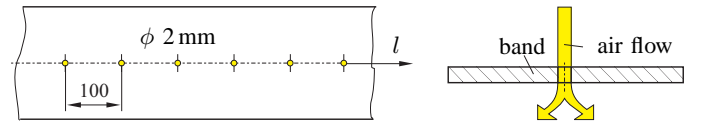}
\caption{AE generation by air flowing through the hole}
\label{trakica3}    
\end{figure}

We performed experiments with two independent continuous AE sources
on an aluminum band of dimensions $4000 \times 40 \times
5\,\text{mm}^3$. Reflections at the end of the band were reduced by
wrapping the ends in putty. The testing area was on the longitudinal
axis in the middle of the band, where 23 holes of diameter 2\,mm and
mutual separation 100\,mm were prepared as shown in Fig. \ref{trakica3}.

\begin{figure}[htb] 
\centering  
\psfrag{preskusanec}[][]{\footnotesize specimen}
\psfrag{senzor}[][]{\footnotesize sensor}
\psfrag{AE sources}[][]{\footnotesize AE source}
\psfrag{ojacevalniki}[][]{\footnotesize amplifier}
\psfrag{A/D pretvorniki}[][]{\footnotesize A/D converter}
\psfrag{lokator}[][]{\footnotesize locator}
\psfrag{risalnik}[][]{\footnotesize plotter}
\psfrag{slepo locevanje}[][]{\footnotesize blind source}
\psfrag{izvorov}[][]{\footnotesize separation}
\includegraphics[width=9cm]{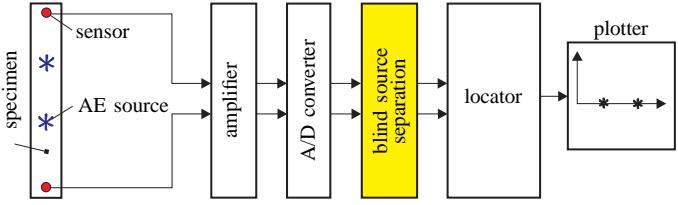} 
\caption{Experimental set-up}
\label{expsetup}
\end{figure}

Two AE sensors were mounted 100\,mm away from the terminal holes, that is
2.4\,m from each other. The origin of the coordinate system was in the
middle of the band and the testing area extended from $-1.1$\,m to
$+1.1$\,m.  AE signals were excited by two independent air jets flowing
through the holes. The source position was arbitrarily selected at $+100$\,mm
and $+800$\,mm. Air jets were formed by two nozzles of diameter
1\,mm using pressure 7 bar. The experimental set-up consisted of the
test specimen (aluminum band), two AE sensors (pinducers), two AE sources (air jets),
two amplifiers, a digital oscilloscope (A/D converter) and a
computer (BSS module, locator, plotter) as shown in Fig.\ \ref{expsetup}. 
Three experiments were performed  : (1) T-D estimation using a
CCF of two AE signals that were not
simultaneously active; (2) T-D estimation using a
CCF of two AE signals which were simultaneously
active and (3) T-D estimation of AE signals using ICA. Location
of sources, based on T-D, by the intelligent locator was
performed in all three cases.

\begin{figure}[htb] 
\centering  
\subfigure[Sensory signal \#1]{
\psfrag{Y}[][]{\small $x_{1}(t)$}
\psfrag{X}[][]{\small $t$ [ms]}
\includegraphics{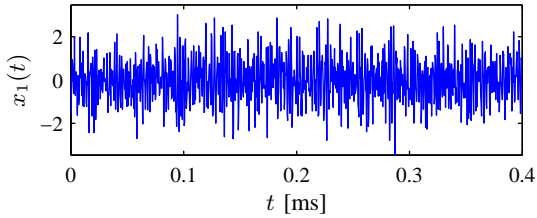}  
\label{zrakhix1} 
}
\subfigure[Sensory signal \#2]{
\psfrag{Y}[][]{\small $x_{2}(t)$}
\psfrag{X}[][]{\small $t$ [ms]}
\includegraphics{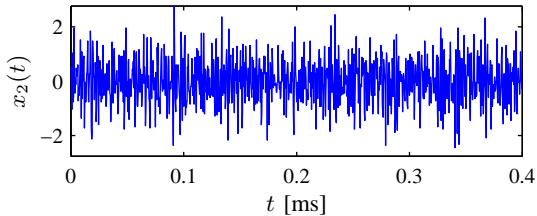} 
\label{zrakhix2}  
}
\caption{Mixtures of two independent continuous AE sources aquired by two sensors}
\label{zrakhix12} 
\end{figure}

\begin{figure}[htb] 
\centering  
\psfrag{X}[][]{\small $n$}
\psfrag{Y}[][]{\small $R_{x_1x_1}$}
\includegraphics{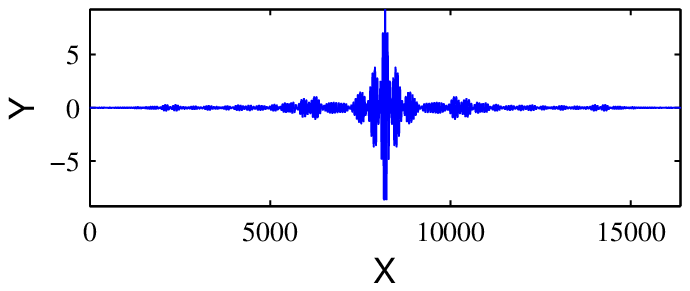}
\psfrag{Y}[][]{\small $R_{x_1x_2}$}
\includegraphics{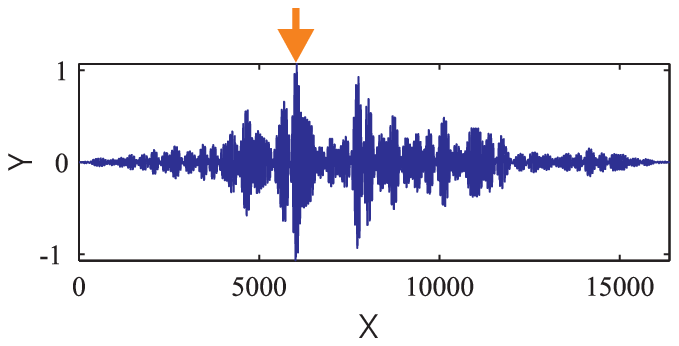}
\psfrag{Y}[][]{\small $R_{x_2x_1}$}
\includegraphics{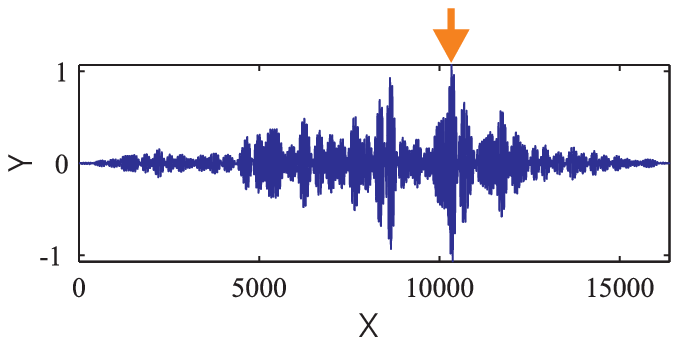}
\psfrag{Y}[][]{\small $R_{x_2x_2}$}
\includegraphics{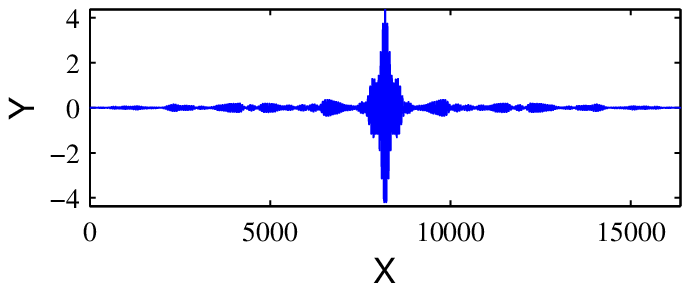}
\caption{Auto- and cross-correlation functions of sensory signals; down-arrow marks the highest peak}
\label{locbsshi_r}    
\end{figure}

In the first experiment only one air jet was activated for a particular
measurement.
In the second experiment both air jets were activated. Sensor signals
were linear convolutive mixtures of two independent continuous AE
sources as shown in Fig. \ref{zrakhix12}. The auto-correlation $R_{11}$, $R_{22}$ and cross-correlation functions
$R_{12}$, $R_{21}$ were calculated from sensor signals. Only one T-D of 
two signals can be estimated from the highest peak in both
CCF, regardless of the number of independent AE
sources on the test specimen as shown in Fig. \ref{locbsshi_r}. This means that a CCF
can not be used for automatic T-D estimation of multiple AE
signals on the test specimen. The CCF exhibits
various peaks which belong to various independent AE sources, but it is ussually
impossible to relate these peaks to corresponding coordinates of AE
sources.

\begin{figure}[htb]
\centering
\psfrag{X}[][]{\small $n$}
\psfrag{Y}[][]{\small ${a}_{11}$}
\includegraphics{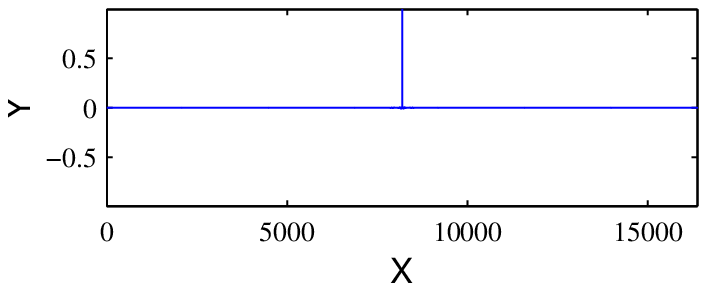}
\psfrag{Y}[][]{\small ${a}_{12}$}
\includegraphics{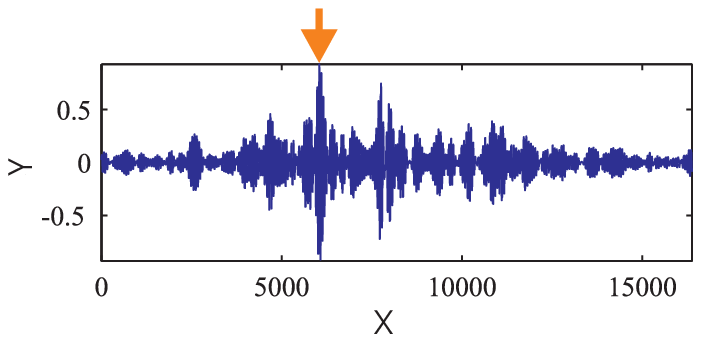}
\psfrag{Y}[][]{\small ${a}_{21}$}
\includegraphics{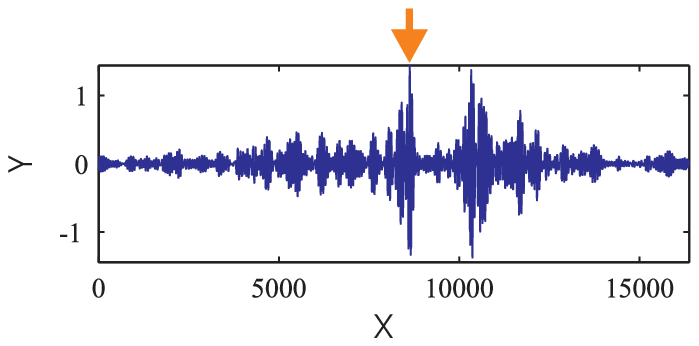}
\psfrag{Y}[][]{\small ${a}_{22}$}
\includegraphics{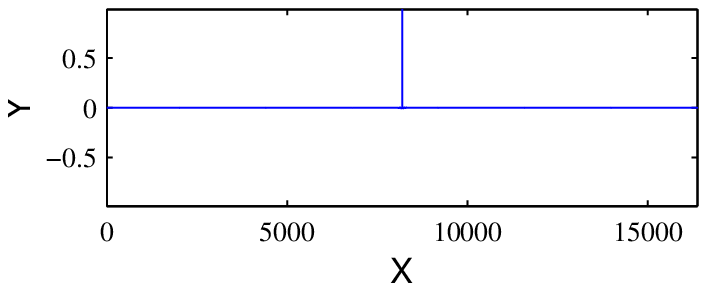}
\caption{Mixing filters obtained by ICA of sensory signals; down-arrow marks the highest peak}
\label{locbsshia}
\end{figure}

In the third experiment the ICA algorithm was used to solve this problem
satisfactorily. The ICA algorithm results in demixing FIR filters which
extract the independent source signals from sensory signals. By
inverting the demixing filters $\vec{W}$ we obtain mixing filters $\vec{A}$.
In the case of two independent AE sources and two sensors, the components
of $\vec{A}$ are four FIR mixing filters, as shown in
Fig.\ \ref{locbsshia}. There are two direct $a_{11}$, $a_{22}$ and two
cross mixing filters $a_{12}$, $a_{21}$. The first index of the filter
represents the number of the sensor, while the second index represents the number of
the source. The position of the highest peak of the cross FIR filters
determines the T-D between two signals from two sensors. If we
substract the coordinate of the highest peak of a direct mixing FIR filter
$a_{11}$ from the coordinate of the highest peak of cross filter $a_{21}$ we
obtain the T-D of first independent AE source, since each of the
highest peaks in the FIR filters belongs to different independent AE
signals. 

\section*{Results}
\begin{figure}[htb]
\centering
\psfrag{X}[][]{\small actual position $l$ [m]}
\psfrag{Y}[][]{\small estimated position $\hat{l}$ [m]}
\psfrag{ICA}[][]{\footnotesize ICA}
\psfrag{orelacijsk}[][]{\footnotesize correlation}
\psfrag{a funkcija}[][]{\footnotesize function}
\psfrag{k}[][]{}
\includegraphics[width=7cm]{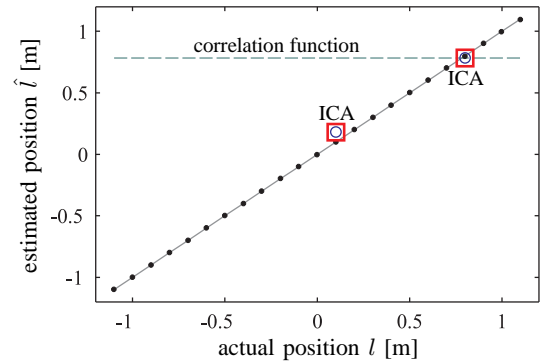}
\caption{Results of location of two continuous independent AE sources.
Symbols: `{\footnotesize $\Box$}' --  AE sources obtained by ICA; `{\Large
$\circ$}' -- estimated AE sources obtained by cross-correlation function
in two steps, when just one of two AE sources was active at time of
measurement;  `$-$ $-$' -- estimated AE sources obtained by cross-correlation
function when two AE sources were active simultaneously; `{\large
$\bullet$}' -- prototype AE sources required for location using intelligent
locator; `$-$' -- distribution of actual sources.}
\label{locbssji}
\end{figure}

The results of T-D estimation of two continuous independent AE
sources are shown in Fig.\ \ref{locbssji}. Three experiments were done.
In the first experiment, the T-D was estimated by a
CCF of two AE sources which were not active
simultaneously as marked by `{\Large $\circ$}'. Locations of these two
sources estimated by the intelligent locator were +181\,mm and +784\,mm.
The second experiment was performed with both AE sources active
simultaneously. T-D were also estimated by a CCF. The highest peak position 
corresponds to the source location marked by `$-$~$-$' and was +784\,mm. 
The third
experiment was performed using ICA for T-D estimation and
location by intelligent locator. The result is marked by `{\footnotesize
$\Box$}'. Estimated positions of this two sources were +179\,mm and
+784\,mm respectively. If we compare the coordinates of both independent AE
sources estimated by the first experiment and by the third experiment, we
find a good correspondence. If we compare estimated AE source coordinates
with actual coordinates, which were +100\,mm and +800\,mm respectively, we
observe a slight disagreement due to experimental error. Experimental
error is about 3\% regarding the distance between sensors. Absolute error
in this case is 79\,mm and 16\,mm respectively. The results also depend
on the number and distribution of prototype sources marked by `{$\bullet$}',
which are essential for operation of the intelligent locator. If the number of
prototype sources is increased, location error is reduced. In our case the
prototype sources were distributed along the beam from $-1.1$\,m to
$+1.1$\,m separated by $0.1$\,m, so that systematic error of the locator
was set to several procents. 

\section*{Discussion and Conclusion}
CCF is applicable to T-D estimation
only in the case of one active AE source. The goal of our research is to
develop a new method to estimate T-D between AE signals in
the case of multiple simultaneously active continuous AE sources. We
have shown that, for this purpose, ICA is
an applicable option. ICA finds a linear coordinate system (the unmixing filters)
such that the resulting signals are statistically independent. This is an advantage of ICA over CCF.   
It represents a new approach to processing of AE data and further
expands the applicability of AE analysis in the field of non-destructive
testing.
In machines or in an industrial environment, multiple sources are usually active
Simultaneously, often representing environmental disturbances. The
corresponding complex signals are not directly applicable to
characterization of particular sources. However, separation of
contributions by ICA analysis in fact represents a kind of filtering,
increasing the applicability of filtered signals to characterization of
sources in complex environments. Future research will be focused on location of multiple AE sources on two-dimensional 
and three-dimensional specimens.

\end{document}